\documentclass{article}

\usepackage{microtype}
\usepackage{graphicx}
\usepackage{subcaption}
\usepackage{booktabs}
\usepackage{multirow}
\usepackage{array}
\usepackage{hyperref}

\usepackage[preprint]{icml2027}
\hypersetup{pdfsubject={arXiv preprint}}

\usepackage{amsmath}
\usepackage{amssymb}
\usepackage{mathtools}
\usepackage{amsthm}
\usepackage[capitalize,noabbrev]{cleveref}

\theoremstyle{plain}

\theoremstyle{definition}

\theoremstyle{remark}

\icmltitlerunning{EpiArb: Hierarchy Residuals for Selective Decision Cascades}

\begin{document}

\twocolumn[
  \icmltitle{EpiArb: Hierarchy Residuals for Selective Cascades \texorpdfstring{\\}{ }in Schema Decision Models}

  \begin{icmlauthorlist}
    \icmlauthor{Shaojie Zhang}{sustech,pcl}
    \icmlauthor{Yue Yu}{pcl}
    \icmlauthor{Yu Zhang}{sustech}
  \end{icmlauthorlist}

  \icmlaffiliation{sustech}{Department of Computer Science and Engineering, Southern University of Science and Technology, Shenzhen, China}
  \icmlaffiliation{pcl}{Peng Cheng Laboratory, Shenzhen, China}
  \icmlcorrespondingauthor{Yu Zhang}{yu.zhang.ust@gmail.com}

  \icmlkeywords{Decision Models, Model Cascading, Selective Inference, Attention Interpretability, Representation Hierarchy}

  \vskip 0.3in
]

\printAffiliationsAndNotice{\textit{Author emails:}
  \href{mailto:12131109@sustech.edu.cn}{12131109@sustech.edu.cn};
  \href{mailto:yuy@pcl.ac.cn}{yuy@pcl.ac.cn}.\quad}

\begin{abstract}
Schema decision models score typed candidate answers in one forward pass, and cascades decide which inputs to escalate to a slower chain-of-thought (CoT) model using the readout's scalar confidence.
Scalar confidence says \emph{whether} the readout is unsure, not \emph{where} the evidence points: when the readout's option pooling and lexical prior wash out a short decisive span, the upstream evidence-routing heads often still prefer the right option.
We measure this head--readout conflict as a \emph{hierarchy residual} and argue that a cascade signal should be judged by how well it ranks inputs by \emph{escalate gain} (fast model wrong, CoT model right), not by how well it detects errors.
By this yardstick the residual's magnitude carries information that confidence lacks, even where the heads do not know the answer; where they do, its direction additionally names a replacement answer and repairs the conflict locally at zero LLM cost.
The residual ranks escalate gain better on every wash-out-prone suite we test, adding it to a learned deferral router lifts accuracy by $2.6$--$5.5$pp on the wash-out-prone public suites, and its ranking advantage, measured on validation before deployment, separates the public suites where it helps from those where Margin suffices.
Under a matched $20\%$ escalate budget to a $35$B CoT model, the residual cascade, with local repair enabled only where validation shows it is safe, reaches $98.3\%$ on two independently seeded multi-field probes, within $0.5$pp of full CoT at one fifth of the calls, and beats logit-margin routing by $+2.5$pp pooled ($N{=}481$, exact McNemar $p{=}0.002$), where routers trained on validation gain labels do no better than Margin.
\end{abstract}

\section{Introduction}
\label{sec:intro}

Production decision pipelines---compliance triage, security audits, evidence checks---need typed answers under tight latency.
Non-autoregressive \emph{decision models}\footnote{Schema-constrained, single-forward classifiers (e.g., Clef-flash, StartLux, Open-Jev), distinct from Decision Transformers in offline RL~\citep{chen2021decision}.} such as Clef-flash~\citep{clefflash2025} score all candidates of a schema in one forward pass, $66\times$ faster than a CoT reasoner in our setup.
The standard way to recover accuracy is a cascade: answer most queries with the fast model and defer the rest to a CoT LLM~\citep{chen2023frugalgpt,dekoninck2025cascade,li2026jevasajudge,reflex2026}.
The deferral rule is usually a scalar such as the logit margin.

Scalar confidence says \emph{whether} the readout is unsure, but not \emph{why}.
In a decision model, a short decisive clause (a revoked credential, a spoofed link, a contradicting sentence) can be outvoted when the readout pools all options into one field state and adds a lexical prior, while the per-option evidence-routing heads upstream of that fusion still align with the clause.
We call this a \textbf{representation hierarchy gap}.
It is measurable: on our security-audit probes the selected heads almost never contradict a correct readout ($3/413$), yet prefer the gold option on $65\%$ of security-policy errors (Table~\ref{tab:gap_evidence}).

A cascade router does not need to detect errors; it needs to rank queries by \emph{escalate gain} (fast model wrong, CoT model right), and the two can disagree.
Where the gap exists, the \emph{magnitude} of head--readout conflict can track gain better than readout uncertainty, even where the heads do not know the answer: they only need to split from the readout more often on gain items.
Where the heads do know the answer, the conflict also has a \emph{direction}: a margin depends on the readout logits alone, so it can flag a query as uncertain but cannot name a replacement, whereas the heads can, which allows a \emph{local} repair with no LLM call.
Ranking is the general mechanism; repair is a special case that is safe only where heads point to gold.

\begin{figure*}[t]
\centering
\includegraphics[width=0.99\linewidth]{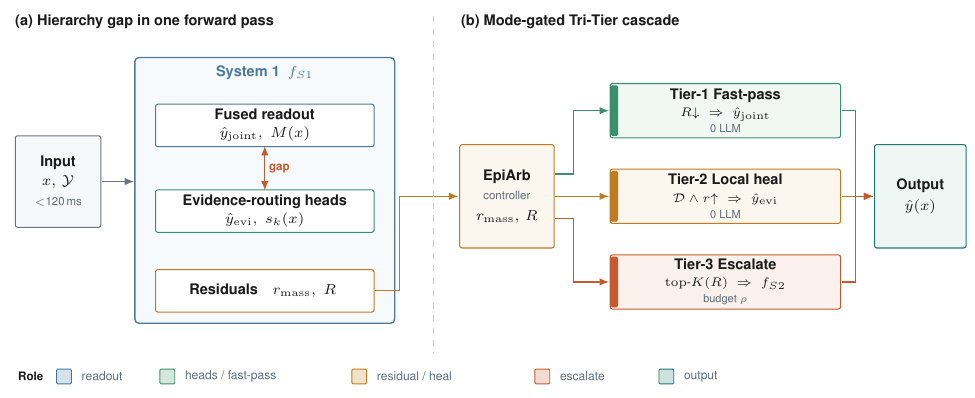}
\caption{\textbf{Selective cascading via EpiArb hierarchy residuals.}
(a)~One System~1 forward pass yields fused readout $\hat{y}_{\mathrm{joint}}$ and evidence scores $s_k(x)$; their mismatch yields residuals $r_{\mathrm{mass}}$ and $R$.
(b)~EpiArb routes each query: low $R$ keeps $\hat{y}_{\mathrm{joint}}$; disagreement $\mathcal{D}$ with high $r_{\mathrm{mass}}$ heals to $\hat{y}_{\mathrm{evi}}$ (0 LLM); else top-$K(R)$ escalates to CoT $f_{S2}$.}
\label{fig:framework}
\end{figure*}

We turn the gap into a cascade signal.
\textbf{EpiArb} (Figure~\ref{fig:framework}) selects a few heads by validation error-AUROC and computes two continuous \emph{hierarchy residuals}: fused alignment-gap risk $R(x)$, which ranks queries for a fixed CoT budget, and runner-up evidence mass $r_{\mathrm{mass}}(x)$, which licenses an optional zero-cost local fix when the heads clearly disagree with the readout.
A validation mode gate enables local fixes only where they heal at least twice as often as they harm; otherwise EpiArb is a pure ranking signal.

The signal is not universally better than a margin, and we characterize when it is.
A three-regime generative model explains why residual ranking beats Margin once a few percent of errors come from wash-out and loses when errors are diffuse.
Its deployable proxy, the ranking advantage measured on validation alone, separates the public suites where the residual wins from those where it only ties Margin, before any test traffic is routed.

\paragraph{Contributions.}
\begin{itemize}
    \item \textbf{A label-free gain signal beyond confidence.} A measurable head--readout gap in schema decision models, and a hierarchy residual whose magnitude ranks queries by escalate gain for a fixed CoT budget; where heads point to gold, its direction also permits zero-cost local repair, enabled by a validation gate only where it is safe (Section~\ref{sec:method}).
    \item \textbf{Why the residual matters.} Evidence that it carries escalate-gain information that scalar confidence lacks: better gain ranking on every wash-out-prone suite, including families where local repair never fires; $+2.6$ to $+5.5$pp when added to a learned deferral router on wash-out-prone public suites; and repair gains that add on top of ranking (Section~\ref{sec:exp_signal}).
    \item \textbf{When it matters, known in advance.} A generative wash-out model whose ranking order flips at $p_{\mathrm{wash}}{\approx}0.05$, and a pre-deployment test built on it: validation-side gain inversion, computed without test data, is favorable on all three wash-out-prone public suites, where the residual wins, and unfavorable on BBH, GSM8K, and ANLI, where it ties Margin (Sections~\ref{sec:method_analysis},~\ref{sec:exp_benchmarks}).
    \item \textbf{Matched-budget evidence.} At $20\%$ escalate to a $35$B CoT model, the cascade reaches $98.3\%$ on two independently seeded audit probes, within $0.5$pp of full CoT; pooled over $N{=}481$ it beats logit margin by $+2.49$pp (McNemar $p{=}0.0018$). On these validation-shifted probes, supervised routers do no better than Margin; on i.i.d.\ public suites they are stronger, which we report as a scope limit (Section~\ref{sec:experiments}).
\end{itemize}

\section{Related Work}
\label{sec:related}

\paragraph{Cascades, routing, and learning to defer.}
Selective prediction abstains when confidence is low~\citep{chow1970optimum,geifman2017selective}; cascades and routers instead hand the input to a stronger model.
Their deferral signals are read from the fast model's \emph{output} or from the query: a learned query--answer score~\citep{chen2023frugalgpt}, token-level uncertainty~\citep{gupta2024language}, calibrated softmax confidence~\citep{guo2017calibration}, isotonic-calibrated aggregate token-margin uncertainty~\citep{ucci2026}, conformal prediction~\citep{vovk2005algorithmic} with set-size deferral~\citep{conformalcascade2026}, answer consistency~\citep{yue2024mot}, self-verification~\citep{aggarwal2024automix}, query-text routing~\citep{ding2024hybrid,ong2025routellm}, or post-output cascade routing~\citep{dekoninck2025cascade}.
Learning-to-defer fits the rule to expert outcomes~\citep{madras2018predict,mozannar2020consistent,narasimhan2022posthoc}, and confidence alone is suboptimal when the expert is a non-uniform specialist~\citep{jitkrittum2023confidence,cortes2026mild}; stronger models are not uniformly better item by item in machine translation and rubric judging~\citep{farinhas2025translate,rao2026wrongplaces}.
On decision models, REFLEX gates on a scalar confidence score~\citep{reflex2026}, while JEV-as-a-Judge gates on the scalar top-label probability~\citep{li2026jevasajudge}.
EpiArb keeps the cascade and changes what it reads: a signal from the decision model's own candidate-conditioned heads, used label-free or as an extra input to the same post-hoc deferral head (LR-wb).

\paragraph{Internal signals for error and hallucination detection.}
Hidden states of autoregressive LLMs encode whether an answer is true~\citep{burns2023ccs,azaria2023internal,orgad2025llmsknow} and can be steered toward truthful outputs~\citep{li2023iti}; NoVo votes using selected attention-head norms at the final sequence position~\citep{novo2025}.
Closest in spirit, Lookback Lens trains a classifier on per-head attention ratios between context and generated tokens~\citep{chuang2024lookback}, and ReDeEP associates RAG hallucinations with excessive parametric-knowledge FFN contribution and insufficient use of retrieved context by copying heads~\citep{sun2025redeep}.
These works share the intuition that internals can disagree with the output, but serve a different setting in three respects.
\emph{Object}: they read a decoder's token-level states, where no head is tied to an answer option; in a schema decision model each candidate $y_k$ issues its own query, so a head yields an evidence score \emph{per option}, directly comparable with the readout's choice.
\emph{Target}: they predict whether an output is wrong; an $S_1{\to}S_2$ cascade needs escalate gain, and the two diverge (\S\ref{sec:exp_signal}).
\emph{Action}: they return a flag; the residual's direction names a replacement option (local repair), and its magnitude allocates a fixed $S_2$ budget.

\paragraph{Layer disagreement and early exit.}
Early-exit models stop at an intermediate layer once its prediction is confident or stable across layers~\citep{xin2020deebert,zhou2020pabee,schuster2022calm}.
They also compare internal and final views of one model, but in the opposite direction: they \emph{save} compute inside the same model when layers agree, whereas EpiArb \emph{buys} accuracy from a different, slower model when heads and readout disagree.

\paragraph{When CoT and pooling fail.}
CoT can improve multi-step reasoning~\citep{wei2022chain,suzgun2022bbh}, with gains concentrated most consistently in math and symbolic tasks~\citep{sprague2025cot}, but can hurt on rule- and exception-driven judgments~\citep{liu2025mindstep}; on PhishNChips our CoT model trails the fast model by $23.7$pp.
Models also use relevant evidence less effectively when it occurs in the middle of long contexts~\citep{liu2024lost}; wash-out is the single-pass, schema-head analogue that the residual measures.
Appendix~\ref{sec:app_related} compares these lines of work along object, signal, target, and action.

\section{Methodology}
\label{sec:method}

\subsection{Problem Setup and Representation Hierarchy}
\label{sec:method_formulation}

Let $x$ be an input context (structured state or document) and $\mathcal{Y}=\{y_1,\ldots,y_C\}$ a finite schema of typed candidate decisions.
A non-autoregressive decision model $f_{S1}$ encodes $(x,\mathcal{Y})$ with a transformer backbone into hidden states $\mathbf{H}\in\mathbb{R}^{T\times d}$ and scores all candidates with a schema head, in a single forward pass.

\paragraph{Schema head and fused readout.}
Each candidate $y_k$ forms its own query from its option span, its lexical embedding, and the question; $L_e$ evidence-routing layers of cross-attention let these per-option queries read $\mathbf{H}$, yielding routed option states $\mathbf{o}_k$.
The readout then fuses them: the $\mathbf{o}_k$ are softmax-pooled into one field state, which decoder layers refine over $\mathbf{H}$ into $\mathbf{f}$, and each candidate is scored as
\begin{equation}
\label{eq:readout}
z(y_k)=\underbrace{\alpha\,\langle\hat{\boldsymbol{\ell}}_k,\hat{\mathbf{a}}\rangle}_{\text{lexical prior}}+\sigma(\gamma)\bigl(\beta\cos(\mathbf{f},\mathbf{o}_k)+g(\mathbf{f},\mathbf{o}_k)\bigr),
\end{equation}
where $\hat{\boldsymbol{\ell}}_k$ is the normalized option embedding, $\hat{\mathbf{a}}$ a question-plus-global anchor, and $g$ an MLP scorer.
This gives prediction $\hat{y}_{\mathrm{joint}}=\arg\max_k z(y_k)$ and logit margin $M(x)=z_{(1)}-z_{(2)}$; let $i^\star$ index $\hat{y}_{\mathrm{joint}}$.
Option pooling averages a localized disqualifying clause with long boilerplate, and the prior encodes strong domain defaults; either can override the clause (\emph{wash-out}).

\paragraph{Directional evidence heads.}
For head $h$ of evidence-routing layer $l$ (Clef-9B: $L_e{=}2$ layers $\times$ $16$ heads) and candidate $y_k$, let query $\mathbf{q}_{l,h}(y_k)$ attend over $\mathbf{H}$ to form attended evidence $\mathbf{c}_{l,h}(y_k)$.
These heads sit upstream of the fusion: they see each option's evidence before it is pooled and mixed with the prior.
Because raw $L_2$ norms $\|\mathbf{c}_{l,h}(y_k)\|_2$ inflate on frequent lexical tokens, we use scale-invariant directional cosine alignment:
\begin{equation}
\label{eq:cos_align}
\mathrm{Cos}_{l,h}(y_k)=\frac{\mathbf{q}_{l,h}(y_k)^\top\mathbf{c}_{l,h}(y_k)}{\|\mathbf{q}_{l,h}(y_k)\|_2\|\mathbf{c}_{l,h}(y_k)\|_2}.
\end{equation}
On a held-out validation split, we score each head by the error-AUROC of its per-head alignment gap for predicting $f_{S1}$ mistakes, selecting the top-$M$ heads $\mathcal{V}$ with signed polarity $\mathrm{dir}_{l,h}\in\{+1,-1\}$.
Aggregated evidence scores are
\begin{equation}
s_k(x)=\frac{1}{|\mathcal{V}|}\sum_{(l,h)\in\mathcal{V}} \mathrm{dir}_{l,h}\,\mathrm{Cos}_{l,h}(y_k),
\end{equation}
with internal evidence prediction $\hat{y}_{\mathrm{evi}}=\arg\max_k s_k(x)$.

\begin{table}[t]
\centering
\small
\caption{\textbf{The hierarchy gap, measured.} Pooled \texttt{Hard}$+$\texttt{Hard\_v2} ($N{=}481$), selected heads frozen on validation. ``Heads$\to$gold'': $\hat{y}_{\mathrm{evi}}$ equals the gold label. Only $3$ of the $68$ errors have above-median margin.}
\label{tab:gap_evidence}
\setlength{\tabcolsep}{3.2pt}
\begin{tabular}{lrrr}
\toprule
\textbf{Family} & \textbf{$S_1$ errors} & \textbf{Heads$\to$gold} & \textbf{Rate} \\
\midrule
Security policy & 23 & 15 & 65\% \\
Verify math & 32 & 5 & 16\% \\
Count / refund & 13 & 1 & 8\% \\
\midrule
All errors & 68 & 21 & 31\% \\
All correct readouts & 413 & 410 & 99\% \\
\bottomrule
\end{tabular}
\end{table}

\paragraph{The gap, measured.}
Table~\ref{tab:gap_evidence} shows the asymmetry EpiArb exploits.
On correct readouts the selected heads almost never disagree ($3/413$), but on $31\%$ of readout errors they prefer the gold option---$65\%$ in security-policy audits, where a buried revocation competes with a benign narrative, and $0$--$16\%$ on arithmetic and counting, where no single span holds the answer.
These errors mostly sit at low margin, so a scalar gate already flags them as uncertain; what it cannot tell is \emph{which} uncertain query a CoT model will actually fix, or, where the heads know the answer, \emph{which} option is right.

\subsection{Continuous Hierarchy Residuals (EpiArb)}
\label{sec:method_epiarb}

EpiArb (\emph{epistemic arbitration}) arbitrates between the fused readout and the upstream evidence.
Binary disagreement $\mathcal{D}(x)=\mathbb{I}(\hat{y}_{\mathrm{joint}}\neq\hat{y}_{\mathrm{evi}})$ is specific to readout errors but too sparse to rank a continuous escalation budget, so we map $\mathbf{s}(x)$ into two continuous residuals.

\paragraph{Beyond a scalar gate.}
Any scalar gate $\phi(z(y_1),\ldots,z(y_C))$, such as margin, max-probability, or entropy, takes the same decision on two inputs with the same readout logits, however their heads disagree.
It can therefore rank escalate gain only insofar as readout uncertainty tracks gain, and it cannot propose a replacement answer.
The residual adds the head axis: its \emph{magnitude} (how far the heads sit from the readout, and how split they are) supports ranking, and, where the heads have read the decisive span, its \emph{direction} ($\hat{y}_{\mathrm{evi}}$ and the mass on it) supports local repair.
The two uses fail differently: a misranked query only wastes one escalate call, while a wrong local flip harms a query the readout got right.
We therefore use a smooth magnitude-based score for ranking and a strict, gated direction-based score for repair.

\paragraph{Escalation ranking residual (Tier-3; magnitude).}
Define the readout alignment gap $g(x)=\max_{k\neq i^\star}s_k(x)-s_{i^\star}(x)$ and internal evidence margin $m_{\mathrm{evi}}(x)=s_{(1)}(x)-s_{(2)}(x)$.
Standardizing each term with validation mean and standard deviation ($z$-scores), the fused residual risk is
\begin{equation}
R(x)=z\bigl(g(x)\bigr)+z\bigl(-m_{\mathrm{evi}}(x)\bigr).
\end{equation}
$R(x)$ is high when the heads contradict the readout (large $g$) or are themselves split (small $m_{\mathrm{evi}}$).

\paragraph{Optional local repair (Tier-2; direction).}
With $\boldsymbol{\pi}(x)=\mathrm{softmax}(\mathbf{s}(x))$, the \emph{runner-up evidence mass}
\begin{equation}
r_{\mathrm{mass}}(x)=1-\pi_{i^\star}(x)
\end{equation}
quantifies internal evidence mass escaping the readout's chosen option $i^\star$.
When $\hat{y}_{\mathrm{evi}}\neq\hat{y}_{\mathrm{joint}}$ and $r_{\mathrm{mass}}(x)\ge\tau_{\mathrm{local}}$ (validation $q_{80}$), Tier-2 overrides $\hat{y}_{\mathrm{joint}}$ with $\hat{y}_{\mathrm{evi}}$ at zero LLM latency.

\paragraph{Mode-gated activation.}
Local flips are only safe where the heads name the gold label (Table~\ref{tab:gap_evidence}).
We therefore fit a mode gate on validation before test: \texttt{ROUTING-only} with $R(x)$ by default, upgraded to \texttt{LOCAL+ROUTING} (L+R) only if heal $\ge 2\times$ harm with no family-level degradation.

\subsection{Mode-Gated Tri-Tier Cascade}
\label{sec:method_tritier}

\begin{algorithm}[tb]
\caption{Matched-budget Tri-Tier Cascade with EpiArb}
\label{alg:cascade}
\begin{algorithmic}[1]
\REQUIRE Query batch $\mathcal{B}$; budget fraction $\rho$; validation-frozen $\mathcal{V}$, $\tau_{\mathrm{local}}$, mode $\in\{\texttt{L+R},\texttt{Route}\}$
\STATE Single forward pass on $f_{S1}$: obtain $\hat{y}_{\mathrm{joint}}$, $\mathbf{s}(x)$, $r_{\mathrm{mass}}(x)$, $R(x)$ for all $x\in\mathcal{B}$
\STATE \textbf{Tier-2 (Local Heal):} if mode $=\texttt{L+R}$, $r_{\mathrm{mass}}(x)\ge\tau_{\mathrm{local}}$, and $\hat{y}_{\mathrm{evi}}\neq\hat{y}_{\mathrm{joint}}$, set $\hat{y}(x)\leftarrow\hat{y}_{\mathrm{evi}}$
\STATE Let $\mathcal{U}\subseteq\mathcal{B}$ be unhealed queries; $K\leftarrow\mathrm{round}(\rho\cdot|\mathcal{B}|)$
\STATE \textbf{Tier-3 (CoT Escalate):} escalate top-$K$ queries in $\mathcal{U}$ ranked by $R(x)$; set $\hat{y}(x)\leftarrow y_{S2}(x)$
\STATE \textbf{Tier-1 (Fast-Pass):} remaining queries in $\mathcal{U}$ return $\hat{y}_{\mathrm{joint}}$
\end{algorithmic}
\end{algorithm}

Algorithm~\ref{alg:cascade} details matched-budget inference.
Errors resolved at Tier-2 leave $\mathcal{U}$ without consuming LLM quota, so all $K$ calls to $f_{S2}$ go to the remaining uncertain instances.

\subsection{When Should Residuals Beat Scalar Confidence?}
\label{sec:method_analysis}

\begin{figure}[t]
\centering
\includegraphics[width=\columnwidth]{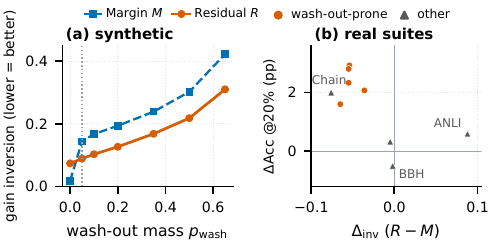}
\caption{\textbf{Residual ranking beats confidence where wash-out occurs.}
(a)~Synthetic three-regime model ($20$ seeds): escalate-gain inversion rate (fraction of gain/non-gain pairs ordered wrongly) of Margin $M$ vs.\ residual $R$; the order flips at $p_{\mathrm{wash}}{=}0.05$.
(b)~Real suites (Table~\ref{tab:applicability}): the measured ranking advantage $\Delta_{\mathrm{inv}}$ against the deployed accuracy gain over Margin at $20\%$ escalate. All five wash-out-prone suites lie in the upper-left quadrant.}
\label{fig:ranking}
\end{figure}

A cascade router does not need to detect errors; it needs to rank queries by \emph{escalate gain}---$S_1$ wrong and $S_2$ right.
We measure this with the gain inversion rate: the fraction of (gain, non-gain) pairs that a risk score orders wrongly.

\paragraph{A generative wash-out model.}
Consider a binary schema $\{\text{prior},\text{gold}\}$ and three regimes.
\emph{Easy} items have a strong local signal $L$ and a correct readout.
\emph{Noise} items have a weak local signal: the readout errs near zero margin and the heads are equally unsure.
\emph{Wash-out} items have a strong local signal that the heads track, but the readout mixes it with a global prior (the analogue of the prior term in Eq.~\ref{eq:readout}), $z=(1{-}\alpha)L+\alpha G$ with $\alpha{=}0.9$, so the readout follows the prior.
We fix the noise mass at $p_{\mathrm{noise}}{=}0.15$, sweep the wash-out mass $p_{\mathrm{wash}}$, use an oracle $S_2$, and compute $M(x)$ and $R(x)$ exactly as above (20 seeds; Appendix~\ref{sec:app_settings}).

\paragraph{Prediction.}
With no wash-out, Margin ranks better (gain inversion $0.018$ vs.\ $0.073$ for $R$): when the heads are as unsure as the readout, they only add noise.
Once wash-out reaches $p_{\mathrm{wash}}{=}0.05$, the order flips ($0.144$ vs.\ $0.088$) and stays flipped up to $p_{\mathrm{wash}}{=}0.65$; the crossing occurs at $0.05$ in $20/20$ seeds (Figure~\ref{fig:ranking}a).
In this model, wash-out errors carry a \emph{confident} wrong readout, so Margin ranks them last, while their heads sit far from the readout, so $R$ ranks them first.
On real data, most wash-out errors already sit at low margin (\S\ref{sec:limitations}), so the residual re-orders the uncertain pool rather than surfacing confident errors, and gains are correspondingly smaller.
The model is a mechanism sketch, not a fitted model: residual routing should win where some errors come from a span the readout overrides, and lose where errors are diffuse.
Since $p_{\mathrm{wash}}$ is unobservable, Section~\ref{sec:exp_benchmarks} tests a deployable proxy: the gain-inversion gap measured on validation.

\section{Experiments}
\label{sec:experiments}

\begin{figure*}[t]
\centering
\includegraphics[width=0.99\linewidth]{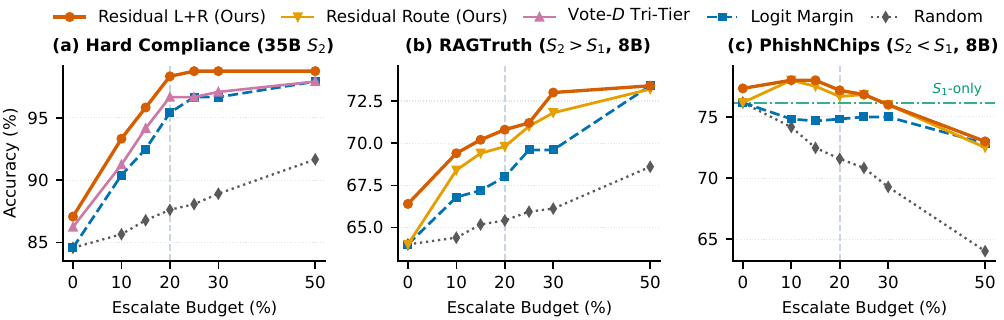}
\caption{\textbf{Matched-budget accuracy curves ($0\%$--$50\%$ escalate).}
(a)~\texttt{Hard} ($35$B): EpiArb reaches $98.33\%$ by $20\%$ and saturates at the $S_2$-only level by $25\%$; Margin needs $50\%$ to reach $97.92\%$.
(b)~\textbf{RAGTruth}: residual dominates Margin and Random up to $30\%$.
(c)~\textbf{PhishNChips} ($S_2{<}S_1$): escalating by Margin or at random falls below $S_1$-only; the residual stays above $S_1$ up to $20\%$.}
\label{fig:pareto}
\end{figure*}

All comparisons use a \textbf{matched escalate budget} on identical System~1 forward caches and identical System~2 CoT decodes.
We report $\rho{=}20\%$ ($5\times$ fewer $S_2$ calls than full CoT); accuracy--budget curves over $0\%$--$50\%$ are in Figure~\ref{fig:pareto}.
We ask: (i)~does the residual close more of the $S_1$--$S_2$ gap than scalar and learned deferral (\S\ref{sec:exp_main})? (ii)~does it carry information that confidence lacks, for repair and for ranking (\S\ref{sec:exp_signal})? (iii)~can we tell \emph{before deployment} where it helps (\S\ref{sec:exp_benchmarks})? (iv)~which design choices matter (\S\ref{sec:exp_ablation})?

\subsection{Experimental Setup}
\label{sec:exp_setup}

\paragraph{Models.}
System~1: Clef-flash 9B~\citep{clefflash2025} as primary, plus StartLux 4B and Open-Jev 0.4B for transfer.
System~2: Qwen3.5-35B-A3B CoT on the multi-field probes and Qwen3-8B CoT on public suites for cost; with the 8B $S_2$, gains also hold on \texttt{Hard} and \texttt{Hard\_v2} (Appendix~\ref{sec:app_hard_v2}).

\paragraph{Controlled multi-field probes.}
White-box cascades need native multi-field schemas and exact single-item error attribution, which public benchmarks do not provide.
We therefore build a suite of $950$ programmatically generated episodes / $1{,}455$ typed questions (Appendix~\ref{sec:app_data}) whose labels are computed in Python from sampled policy or arithmetic predicates (fixed seeds; not LLM-authored).
Splits: fit split \texttt{val} ($N{=}824$); test splits \texttt{Hard} ($N{=}240$) and its independently seeded replicate \texttt{Hard\_v2} ($N{=}241$), where $S_1$ errors are dominated by span wash-out; and \texttt{Chain} ($N{=}150$), where they are dominated by multi-hop compute depth.
All EpiArb parameters ($\mathcal{V}$, $z$-scores, $\tau_{\mathrm{local}}$, mode gate) are fit on \texttt{val} and frozen before any test split is scored.

\paragraph{Public suites.}
Wash-out-prone: RAGTruth~\citep{niu2024ragtruth} ($N{=}500$), PhishNChips~\citep{phishnchips2026} ($N{=}600$), and Clickbait~\citep{chakraborty2016clickbait} ($N{=}500$).
Scope controls: BBH~\citep{suzgun2022bbh} ($6$ tasks), GSM8K~\citep{cobbe2021gsm8k}, ANLI~\citep{nie2020anli}.
For each suite, residual parameters are fit once on a held-out validation split and frozen before the test split.

\paragraph{Baselines.}
Under the same $\rho$: Random ($10$ seeds); Max-Softmax / temperature-scaled confidence~\citep{guo2017calibration}; Predictive Entropy~\citep{gupta2024language}; Logit Margin $M(x)$ and its isotonic calibration; \textbf{LR-bb}, post-hoc learning-to-defer~\citep{jitkrittum2023confidence,narasimhan2022posthoc} (logistic regression on margin/confidence/entropy, trained to predict escalate gain); \textbf{LR-wb}, the same head given residual features $(R,r_{\mathrm{mass}},g,m_{\mathrm{evi}},\mathcal{D})$; a \textbf{hidden-state linear probe}~\citep{azaria2023internal} on the full pooled, question, and per-field states of $S_1$ (PCA-$128$, $L_2$-regularized logistic regression, trained on validation to predict escalate gain); two NoVo~\citep{novo2025} adaptations, \textbf{NoVo$_V$} (late-half $L_2$ majority vote as answer, $0$ LLM calls) and \textbf{NoVo$_E$} (escalate by norm disagreement); and binary \textbf{Vote-$D$} Tri-Tier, which uses the same heads but a binary disagreement flag instead of the continuous residual.
All learned baselines see validation $S_2$ outcomes; EpiArb does not use them for ranking.

\begin{table*}[t]
\centering
\small
\caption{\textbf{Primary multi-field probes under a matched $20\%$ escalate budget (Clef-9B $\to$ Qwen3.5-35B CoT).}
Identical $S_1$ states and $S_2$ CoT caches. ``Gap closed'' $=(\mathrm{Acc}-S_1)/(S_2-S_1)$.
\textbf{Ours} is mode-gated on validation: L+R on \texttt{Hard}, Route on \texttt{Chain}; Combined pools the $20\%$ budget over $N{=}390$.
Single-stream A100 latency: $S_1$ $0.14$s, $S_2$ $9.29$s, any $20\%$ cascade $2.00$s ($4.6\times$ faster than full CoT).}
\label{tab:main_results}
\setlength{\tabcolsep}{5.0pt}
\begin{tabular}{lccccc}
\toprule
& \multicolumn{3}{c}{\textbf{Accuracy (\%)}} & \multicolumn{2}{c}{\textbf{Gap closed (\%)}} \\
\cmidrule(lr){2-4} \cmidrule(lr){5-6}
\textbf{System / Controller} & \textbf{Hard ($N{=}240$)} & \textbf{Chain ($N{=}150$)} & \textbf{Combined ($N{=}390$)} & \textbf{Hard} & \textbf{Combined} \\
\midrule
$S_1$ only (Clef-9B, $0\%$ esc.) & 84.58 & 45.33 & 69.49 & 0 & 0 \\
$S_2$ only (35B CoT, $100\%$ esc.) & 98.75 & 98.00 & 98.46 & 100 & 100 \\
Oracle ($S_1 \lor S_2$) & 99.58 & 98.00 & 98.97 & --- & --- \\
\midrule
\multicolumn{6}{l}{\textit{$0\%$ escalate}} \\
NoVo$_V$ (late-half vote) & 86.67 & 28.67 & 64.36 & 15 & $-18$ \\
Residual Local only & 87.08 & 52.00 & 73.59 & 18 & 14 \\
\midrule
\multicolumn{6}{l}{\textit{$20\%$ escalate}} \\
Random (10-seed mean) & 87.63 & 55.73 & 75.59 & 22 & 21 \\
Predictive Entropy & 93.33 & 59.33 & 83.08 & 62 & 47 \\
Isotonic Margin & 95.00 & 59.33 & 83.08 & 74 & 47 \\
Max-Conf / Temp-Conf & 95.42 & 60.00 & 82.82 & 76 & 46 \\
Logit Margin & 95.42 & 60.00 & 82.56 & 76 & 45 \\
LR-bb (black-box deferral) & 95.42 & 60.00 & 82.82 & 76 & 46 \\
LR-wb (residual features) & 93.33 & 58.00 & 80.26 & 62 & 37 \\
Hidden-state probe & 95.42 & \textbf{62.67} & 82.31 & 76 & 44 \\
NoVo$_E$ (norm disagreement) & 92.92 & 61.33 & 84.10 & 59 & 50 \\
Vote-$D$ Tri-Tier (binary) & 96.67 & \textbf{62.67} & 84.36 & 85 & 51 \\
\textbf{EpiArb residual (ours)} & \textbf{98.33} & 62.00 & \textbf{85.90} & \textbf{97} & \textbf{57} \\
\bottomrule
\end{tabular}
\end{table*}

\subsection{Primary Probes: Accuracy, Replication, and Decomposition}
\label{sec:exp_main}

\paragraph{\texttt{Hard}.}
Every scalar gate plateaus at or below $95.42\%$ (Table~\ref{tab:main_results}).
Learned routers do not escape it (LR-bb and the probe tie Margin; LR-wb drops to $93.33\%$): \texttt{Hard} is shifted from validation, and routers fit to validation gain labels do not transfer, whereas the residual's ranking uses no $S_2$ labels (\S\ref{sec:exp_benchmarks}).
EpiArb reaches \textbf{$98.33\%$}, within $0.42$pp of full CoT at one fifth of the $S_2$ calls, closing $97\%$ of the $S_1$--$S_2$ gap versus $76\%$ for Margin ($+2.92$pp, paired-bootstrap $95\%$ CI $[+0.83,+5.42]$, exact McNemar $p{=}0.039$, $8$ vs.\ $1$ discordant).
Against binary Vote-$D$ the gain is $+1.67$pp (CI $[0,+3.75]$, $p{=}0.22$): the continuous residual is consistently better, but on one $N{=}240$ split this is not significant.

\paragraph{\texttt{Chain} (a compute-depth regime).}
On \texttt{Chain}, $S_1$ errors stem from multi-hop arithmetic, so the heads rarely ``know'' the answer and the validation gate selects Route.
Residual Route reaches $62.00\%$ ($+2.00$pp vs.\ Margin, CI $[-4.7,+8.7]$); every ranking signal closes only about a third of a $53$pp gap that a $20\%$ budget cannot cover.
Unconstrained L+R would reach $66.00\%$, so the conservative gate costs accuracy here; we keep it because it needs no test labels and protects Open-Jev-0.4B (\S\ref{sec:exp_ablation}).

\begin{table}[t]
\centering
\small
\caption{\textbf{Replication and pooled significance.}
Frozen recipe on independently seeded \texttt{Hard\_v2} ($N{=}241$) and pooled with \texttt{Hard} ($N{=}481$), matched $20\%$ to 35B CoT.
$p$: exact McNemar vs.\ the row's comparator on pooled items.}
\label{tab:hard_v2}
\setlength{\tabcolsep}{3.0pt}
\resizebox{\columnwidth}{!}{%
\begin{tabular}{lccc}
\toprule
\textbf{System} & \textbf{Hard\_v2} & \textbf{Pooled} & \textbf{Ours $-$ row (pooled)} \\
\midrule
$S_1$ only & 87.14 & 85.86 & --- \\
$S_2$ only ($100\%$) & 98.34 & 98.54 & --- \\
Residual Local ($0\%$) & 90.46 & 88.77 & --- \\
Logit Margin & 96.27 & 95.84 & $\mathbf{+2.49}$ ($p{=}0.0018$) \\
Vote-$D$ Tri-Tier & 97.93 & 97.30 & $+1.04$ ($p{=}0.125$) \\
\textbf{EpiArb residual} & \textbf{98.34} & \textbf{98.34} & --- \\
\bottomrule
\end{tabular}%
}
\end{table}

\paragraph{Replication and pooled significance (primary claim).}
On independently seeded \texttt{Hard\_v2}, the frozen recipe matches full CoT ($98.34\%$ vs.\ $98.34\%$) at $20\%$ escalation; zero-cost local healing alone gains $+3.32$pp over $S_1$ ($8$ heals~/~$0$ harms, $p{=}0.0078$).
Pooling the two non-overlapping test splits (Table~\ref{tab:hard_v2}), EpiArb beats Margin by $+2.49$pp (CI $[+1.04,+4.16]$; $b{=}13$, $c{=}1$; $\mathbf{p{=}0.0018}$).
Against binary Vote-$D$ the pooled gain is $+1.04$pp ($p{=}0.125$): the residual's advantage over scalar confidence is significant, while its advantage over a binary use of the same heads is directionally consistent but not.

\begin{table}[t]
\centering
\small
\caption{\textbf{Decomposition on \texttt{Hard}} (matched $20\%$).
Left: swapping each tier into a Margin cascade. Right: zero-cost local $\Delta$pp over $S_1$ by family.}
\label{tab:knockout}
\setlength{\tabcolsep}{3.0pt}
\resizebox{\columnwidth}{!}{%
\begin{tabular}{lc|lcc}
\toprule
\textbf{Cascade (@20\%)} & \textbf{Acc} & \textbf{Family (0 LLM)} & \textbf{Res.} & \textbf{Vote-$D$} \\
\midrule
Margin only & 95.42 & Security ($n{=}62$) & $\mathbf{+9.68}$ & $+6.45$ \\
Res.\ Local $+$ Margin & 96.67 & Verify math ($n{=}36$) & $0.00$ & $0.00$ \\
Res.\ Route only & 97.08 & Refund chain ($n{=}80$) & $0.00$ & $0.00$ \\
Vote-$D$ Local $+$ Res.\ Route & 97.50 & SLA / Count ($n{=}62$) & $0.00$ & $0.00$ \\
\textbf{Res.\ Local $+$ Res.\ Route} & \textbf{98.33} & \textbf{Overall} & $\mathbf{+2.50}$ & $+1.67$ \\
\bottomrule
\end{tabular}%
}
\end{table}

\paragraph{Ranking carries the larger share; repair adds on top.}
Table~\ref{tab:knockout}: relative to a Margin cascade, residual routing alone adds $+1.67$pp and residual local healing $+1.25$pp; together they add $+2.92$pp, i.e.\ the two tiers are additive---healing frees escalate quota that routing spends on the next-highest-gain items.
Local heals are confined to the security family ($+9.68$pp; $+18.2$pp on \texttt{Hard\_v2}), exactly where Table~\ref{tab:gap_evidence} shows heads most often point to gold; families where heads rarely know the answer see zero flips and zero harms.

\begin{table*}[t]
\centering
\small
\caption{\textbf{Public suites (Clef-9B $\to$ Qwen3-8B CoT, matched $20\%$).}
All learned and residual parameters fit once on a held-out validation split.
LR-bb, LR-wb, and the hidden-state probe are trained on validation escalate-gain labels; residual Route/L+R are not.}
\label{tab:public_benchmarks}
\setlength{\tabcolsep}{2.4pt}
\begin{tabular}{llcccccccccc}
\toprule
& & & & \multicolumn{3}{c}{\textbf{Scalar / vote}} & \multicolumn{3}{c}{\textbf{Learned deferral}} & \multicolumn{2}{c}{\textbf{EpiArb residual}} \\
\cmidrule(lr){5-7} \cmidrule(lr){8-10} \cmidrule(lr){11-12}
\textbf{Suite} & \textbf{$N$} & \textbf{$S_1$} & \textbf{$S_2$} & \textbf{Margin} & \textbf{NoVo$_V$} & \textbf{NoVo$_E$} & \textbf{LR-bb} & \textbf{LR-wb} & \textbf{Probe} & \textbf{Route} & \textbf{L+R} \\
\midrule
\multicolumn{12}{l}{\textit{Wash-out-prone}} \\
RAGTruth & 500 & 64.0 & 71.2 & 68.0 & 45.6 & 68.8 & 68.6 & 71.8 & \textbf{74.8} & 69.8 & 70.8 \\
PhishNChips & 600 & 76.2 & 52.5 & 74.8 & 23.8 & 73.3 & 75.0 & \textbf{80.5} & 80.3 & 76.7 & 77.2 \\
Clickbait & 500 & 87.0 & 82.0 & 88.4 & 90.8 & 90.0 & 88.4 & \textbf{91.0} & 89.8 & 90.0 & 90.0 \\
\midrule
\multicolumn{12}{l}{\textit{Scope controls}} \\
BBH ($6$-task) & 600 & 63.8 & 63.8 & 70.3 & 50.8 & 65.2 & 69.3 & 70.2 & \textbf{71.0} & 69.8 & 69.3 \\
GSM8K & 300 & 81.0 & 69.3 & 80.7 & 28.3 & 76.7 & 80.3 & 80.7 & 80.7 & \textbf{81.0} & 80.0 \\
ANLI & 500 & 55.0 & 63.8 & 57.2 & 20.6 & 55.6 & 57.0 & 56.0 & 56.0 & \textbf{57.8} & 56.2 \\
\bottomrule
\end{tabular}
\end{table*}

\subsection{Why the Residual Matters: Two Roles, Two Preconditions}
\label{sec:exp_signal}

The accuracy gains above could in principle come from a better-tuned threshold rather than from new information.
We therefore test the residual as a \emph{signal}, separately for its two jobs, without the cascade around it (Table~\ref{tab:applicability}, Figure~\ref{fig:ranking}).

\begin{table}[t]
\centering
\small
\caption{\textbf{The residual as a ranking signal, and where it helps.}
$\Delta_{\mathrm{inv}}$: escalate-gain inversion rate of $R$ minus Margin ($={-}\Delta$gain-AUROC; negative $=$ residual ranks gain items better), cross-fit inside validation (val; no test data, Appendix~\ref{sec:app_data}) or measured on test.
$^\dagger$The three probes share one validation split.
$\Delta$Acc: deployed residual minus Margin @$20\%$ (pp), paired-bootstrap $95\%$ CI.
$+$Res.: accuracy change when residual features are added to the same learned deferral router (LR-wb $-$ LR-bb, pp); negative on the shifted probes, where validation-fit routers overfit (\S\ref{sec:exp_main}).}
\label{tab:applicability}
\setlength{\tabcolsep}{2.6pt}
\begin{tabular}{lccccc}
\toprule
& \multicolumn{2}{c}{\textbf{$\Delta_{\mathrm{inv}}$}} & & & \\
\cmidrule(lr){2-3}
\textbf{Suite} & \textbf{val} & \textbf{test} & \textbf{$\Delta$Acc} & \textbf{95\% CI} & \textbf{$+$Res.} \\
\midrule
\multicolumn{6}{l}{\textit{Wash-out-prone}} \\
\texttt{Hard} & $+0.066^\dagger$ & $-0.054$ & $+2.92$ & $[+0.8,+5.4]$ & $-2.1$ \\
\texttt{Hard\_v2} & $+0.066^\dagger$ & $-0.036$ & $+2.07$ & $[+0.4,+4.1]$ & --- \\
RAGTruth & $-0.047$ & $-0.055$ & $+2.80$ & $[+0.8,+5.0]$ & $+3.2$ \\
PhishNChips & $-0.041$ & $-0.055$ & $+2.33$ & $[+1.0,+3.8]$ & $+5.5$ \\
Clickbait & $-0.049$ & $-0.065$ & $+1.60$ & $[+0.2,+3.0]$ & $+2.6$ \\
\midrule
\multicolumn{6}{l}{\textit{Compute depth / reasoning}} \\
\texttt{Chain} & $+0.066^\dagger$ & $-0.076$ & $+2.00$ & $[-4.7,+8.7]$ & $-2.0$ \\
BBH & $+0.022$ & $-0.002$ & $-0.50$ & tie & $+0.9$ \\
GSM8K & $+0.032$ & $-0.005$ & $+0.33$ & tie & $+0.4$ \\
ANLI & $+0.114$ & $+0.088$ & $+0.60$ & tie & $-1.0$ \\
\bottomrule
\end{tabular}
\end{table}

\paragraph{Ranking does not need heads that know the answer.}
On \texttt{Hard}, local repair fires only in the security family, where heads point to gold (Table~\ref{tab:gap_evidence}), yet the ranking advantage is \emph{larger} outside it. Over all of \texttt{Hard}, $R$ puts $32$ of the $36$ gain items in its top $20\%$ versus $28$ for Margin; gain AUROC rises from $0.887$ to $0.956$ on the non-security families, where heads name the gold option on at most $16\%$ of errors, versus $0.930$ to $0.956$ on security.
On \texttt{Chain}, where local repair is gated off entirely, $R$ still ranks gain better ($0.759\to0.836$).
Direction needs heads that are right; magnitude needs only heads that split from the readout more often on gain items.
Because the two jobs act on different items, their gains add with zero interaction (Table~\ref{tab:knockout}).

\paragraph{Gain, not error.}
On \texttt{Hard}'s validation split, Margin predicts $S_1$ \emph{errors} better than $R$ (AUROC $0.893$ vs.\ $0.841$), yet $R$ predicts escalate \emph{gain} better on test ($0.957$ vs.\ $0.903$): error AUROC is the wrong yardstick for a cascade signal.

\paragraph{The residual adds information beyond black-box confidence.}
If the residual were a re-encoding of confidence, adding it to a learned router would not help.
On the i.i.d.\ public suites, the same logistic deferral head gains $+3.2$, $+5.5$, and $+2.6$pp on RAGTruth, PhishNChips, and Clickbait when residual features are added to margin, max-probability, and entropy, and changes by at most $1$pp on BBH, GSM8K, and ANLI, where the residual carries no extra gain signal.
For local repair, zeroing the cosine features of the eight selected heads removes $57\%$ of beneficial heals and injecting Gaussian noise into them removes $86\%$ (Appendix~\ref{sec:app_causal}).

\subsection{Public Suites: Can We Tell in Advance?}
\label{sec:exp_benchmarks}

Section~\ref{sec:method_analysis} suggests that the residual helps when it ranks escalate-gain items better than Margin, i.e.\ when $\Delta_{\mathrm{inv}}{=}\mathrm{inv}_R{-}\mathrm{inv}_M{<}0$.
Table~\ref{tab:applicability} reports this quantity on test and, without any test data, on validation; Table~\ref{tab:public_benchmarks} gives the full accuracies.

\paragraph{Ranking advantage and accuracy gain agree.}
On all five wash-out-prone suites the residual ranks gain items better on test ($\Delta_{\mathrm{inv}}\approx-0.05$) and wins by $+1.6$ to $+2.9$pp, with every bootstrap CI excluding zero; on BBH, GSM8K, and ANLI, where errors are reasoning failures rather than buried evidence, $\Delta_{\mathrm{inv}}\gtrsim0$ and accuracy ties (Figure~\ref{fig:ranking}b).

\paragraph{A pre-deployment test on public suites.}
Measured on validation alone, $\Delta_{\mathrm{inv}}$ is negative on RAGTruth, PhishNChips, and Clickbait ($-0.041$ to $-0.049$), where the residual wins, and positive on BBH, GSM8K, and ANLI ($+0.022$ to $+0.114$), where it ties Margin.
Deploying the residual only where validation $\Delta_{\mathrm{inv}}{<}0$ thus keeps every public gain and falls back to Margin where it is as good.
Under shift the test is conservative: the primary probes' validation holds few wash-out errors ($\Delta_{\mathrm{inv}}{=}{+}0.066$), yet the residual wins on test; it can miss a gain but never approves a loss.

\paragraph{Supervised routers under i.i.d.\ vs.\ shifted test data.}
On i.i.d.\ public suites, gain-label routers are strong: the hidden-state probe beats the residual cascade on RAGTruth ($+4.0$pp) and PhishNChips ($+3.1$pp), as does LR-wb; on the shifted probes it only ties Margin.
The residual is thus a label-free, shift-robust signal and a router input (Table~\ref{tab:applicability}), not a substitute for in-distribution gain labels.

\subsection{Transfer and Design Ablations}
\label{sec:exp_ablation}

\paragraph{Transfer.}
With frozen residual definitions, StartLux-4B reproduces the $+2.50$pp local lift on \texttt{Hard}; on Open-Jev-0.4B, local flips would cost $29$pp on validation counting, so the gate selects Route ($+1.25$pp on \texttt{Hard}, $+5.20$pp on RAGTruth; Appendix Table~\ref{tab:cross_model}).

\paragraph{Selection, continuity, stability.}
Random eight-head ensembles lose $7.50$pp and all heads $7.92$pp; the continuous mass residual heals more than a binary flag ($+2.50$ vs.\ $+1.67$pp); over $10$ validation refits, local gain is $+2.50\pm0.00$pp and routing gain $+1.37\pm0.28$pp (Appendix Table~\ref{tab:residual_ctrl_ablation}).

\section{Discussion and Limitations}
\label{sec:discussion}

\paragraph{What the residual adds beyond confidence.}
Confidence is a property of the readout; the residual measures how far the readout has drifted from evidence the model itself computed.
Ranking asks only whether the heads \emph{disagree} with the readout; local repair additionally asks whether they \emph{know} the answer.
Where errors are computations (\texttt{Chain}, BBH, GSM8K) or whole-premise inferences (ANLI), Margin remains the better default, as validation $\Delta_{\mathrm{inv}}$ flags in advance.

\subsection{Limitations}
\label{sec:limitations}

\textbf{(1)~Wash-out errors have low margin too.}
On \texttt{Hard}$+$\texttt{Hard\_v2}, only $3$ of $68$ errors have above-median margin: residuals do not find errors that Margin cannot see at all; they re-order the low-margin pool by escalate gain.
Gains are thus moderate ($+2$--$3$pp), and not significant against binary Vote-$D$ ($p{=}0.125$ pooled).

\textbf{(2)~Local healing is family-specific.}
Zero-cost flips concentrate in the security family; on \texttt{Chain} and Open-Jev-0.4B the gate disables healing, which is safe but costs $4$pp on \texttt{Chain}.

\textbf{(3)~Scope of the evidence.}
\texttt{Hard}/\texttt{Hard\_v2}/\texttt{Chain} are generated with exact labels (Appendix~\ref{sec:app_data}), not organic logs; with i.i.d.\ gain labels, supervised routers are stronger (\S\ref{sec:exp_benchmarks}).

\textbf{(4)~Serving.}
The $4.6\times$ speedup assumes single-stream $S_2$; under batching the robust claim is $5\times$ fewer $S_2$ calls. Open-ended outputs would need candidate clustering.

\section{Conclusion}
\label{sec:conclusion}

Schema decision models err when the fused readout washes out a short decisive span that upstream evidence-routing heads still encode.
A hierarchy residual measures this conflict in the same forward pass: its magnitude out-ranks confidence by escalate gain, and its direction repairs conflicts at zero cost.
At $20\%$ budget it comes within $0.5$pp of full CoT and beats Margin by $+2.5$pp ($p{=}0.002$); a validation test flags where it does not help.

\clearpage
\section*{Impact Statement}
\label{sec:impact}

This work targets structured decision pipelines whose failures concentrate when pooling washes out localized evidence---common in compliance and risk triage.
Compact single-forward models handle routine traffic; hierarchy residuals escalate only queries with elevated head--readout conflict.
At a $20\%$ escalate budget this cuts CoT calls by roughly $5\times$.
Deployments should apply the validation mode gate and follow the applicability map (Table~\ref{tab:applicability}).

\bibliography{references}
\bibliographystyle{icml2027}

\newpage
\appendix
\onecolumn
\section{Extended Related Work}
\label{sec:app_related}

Table~\ref{tab:app_related} places EpiArb among work that either defers from a fast model to a slower one or reads model internals to judge an output.
Almost any escalation method can be compared with any other, so we compare along the four axes that define our setting rather than by task.
\emph{Object}: which model the signal is read from.
\emph{Signal}: what is read.
\emph{Target}: what the signal is trained or designed to predict.
\emph{Action}: what is done with it.

\begin{table}[ht]
\centering
\small
\caption{\textbf{Positioning by object, signal, target, and action.} ``AR LLM'': autoregressive language model. ``Decision model'': schema-constrained, single-forward model scoring typed candidates. $S_2$: the slower model receiving escalated inputs.}
\label{tab:app_related}
\setlength{\tabcolsep}{3.2pt}
\begin{tabular}{>{\raggedright\arraybackslash}p{3.5cm}>{\raggedright\arraybackslash}p{2.1cm}>{\raggedright\arraybackslash}p{3.6cm}>{\raggedright\arraybackslash}p{2.9cm}>{\raggedright\arraybackslash}p{3.6cm}}
\toprule
\textbf{Work} & \textbf{Object} & \textbf{Signal} & \textbf{Target} & \textbf{Action} \\
\midrule
FrugalGPT~\citep{chen2023frugalgpt}; LM cascades~\citep{gupta2024language} & AR LLM & Learned query--answer score / token uncertainty & Answer acceptable & Stop or defer \\
MoT cascade~\citep{yue2024mot}; AutoMix~\citep{aggarwal2024automix} & AR LLM & Answer consistency / self-verification & Answer correct & Stop or defer \\
HybridLLM~\citep{ding2024hybrid}; RouteLLM~\citep{ong2025routellm} & Query only & Router on query text & Quality gap / preference & Pick a model before answering \\
Cascade routing~\citep{dekoninck2025cascade} & AR LLM & Query and post-hoc output features & Quality / cost trade-off & Route or continue after outputs \\
UCCI~\citep{ucci2026}; Conformal cascade~\citep{conformalcascade2026} & AR LLM & Calibrated token-margin uncertainty / conformal set & Error / coverage & Defer at a cost or risk level \\
REFLEX~\citep{reflex2026}; JEV-as-a-Judge~\citep{li2026jevasajudge} & Decision model & Scalar confidence / top-label probability & Error / non-inferiority & Defer to $S_2$ \\
Post-hoc L2D~\citep{narasimhan2022posthoc,jitkrittum2023confidence} & Any & Output features & Expert-outcome-aware gain & Defer \\
MILD~\citep{cortes2026mild} & Any & Input features and expert correctness/cost & Cost-sensitive deferral & Defer \\
\midrule
CCS~\citep{burns2023ccs}; SAPLMA-style probes~\citep{azaria2023internal}; \citet{orgad2025llmsknow} & AR LLM & Hidden states & Truthfulness / error & Answer binary questions / flag \\
ITI~\citep{li2023iti} & AR LLM & Truth directions in heads & Truthfulness & Steer generation \\
NoVo~\citep{novo2025} & AR LLM & Selected head norms at the final token & Answer choice & Replace answer by vote \\
Lookback Lens~\citep{chuang2024lookback} & AR LLM & Context-vs-generated attention ratio per head & Contextual hallucination & Flag; guided decoding \\
ReDeEP~\citep{sun2025redeep} & AR LLM (RAG) & Copying-head vs.\ knowledge-FFN contribution & RAG hallucination & Flag; reweight modules \\
DeeBERT / PABEE / CALM~\citep{xin2020deebert,zhou2020pabee,schuster2022calm} & Encoder / AR LM & Intermediate-layer confidence or agreement & Safe to exit early & Exit with fewer layers \\
\midrule
\textbf{EpiArb (ours)} & \textbf{Decision model} & \textbf{Per-candidate head evidence vs.\ pooled readout} & \textbf{Escalate gain ($S_1$ wrong, $S_2$ right)} & \textbf{Rank a fixed $S_2$ budget; repair locally to a named option} \\
\bottomrule
\end{tabular}
\end{table}

\paragraph{Cascades on decision models.}
REFLEX~\citep{reflex2026} returns a typed Jev decision and its confidence, executes high-confidence decisions, and hands the rest to a strong LLM. JEV-as-a-Judge~\citep{li2026jevasajudge} accepts when the top-label probability exceeds a threshold frozen for non-inferiority against the fallback judge.
Both establish the $S_1{\to}S_2$ product form we build on, and both use a black-box scalar.
Our question is complementary: given the same cascade, can the decision model's internals tell \emph{which} low-confidence inputs the slower model will actually fix, and sometimes fix them itself?
On the tested rubric-judging panels, decision models and LLMs are often wrong in the same places~\citep{rao2026wrongplaces}; this is one reason a gain target, rather than an error target, matters for the router.

\paragraph{Why internal-signal detectors do not transfer directly.}
Detectors such as Lookback Lens and ReDeEP are defined over generated tokens: Lookback Lens needs a span of new tokens to form its context-versus-generated ratio, and ReDeEP measures how much copying heads and FFNs contribute to each generated token.
A decision model generates no tokens.
Its candidates are scored in one pass, and each candidate's query attends over the context, so a head yields a per-option evidence vector $\mathrm{Cos}_{l,h}(y_k)$ that can be compared with the readout's own ranking of the same options (\S\ref{sec:method_formulation}).
The hierarchy residual is this comparison; it has no counterpart in token-level detectors, which is also why its direction can propose a replacement answer.
The hidden-state probe in our baselines is the closest transferable analogue of the probing line~\citep{azaria2023internal,orgad2025llmsknow}: it is strong when validation and test match, and ties Margin on the shifted probes (\S\ref{sec:exp_benchmarks}).

\paragraph{Gain versus error.}
Most internal-signal work, and most confidence gates, target $S_1$ error.
Post-hoc learning-to-defer~\citep{narasimhan2022posthoc,jitkrittum2023confidence,mozannar2020consistent} targets the expected improvement from deferring, but needs $S_2$ outcomes on a labeled set.
The residual's ranking is fit without $S_2$ labels (only head selection and two $z$-scores), and we evaluate it by gain inversion, the quantity a fixed-budget cascade actually optimizes.

\section{Implementation, Data Protocols, and Extended Results}
\label{sec:appendix}

\subsection{Model Specifications and Compute}
\label{sec:app_hyperparams}

\begin{table}[ht]
\centering
\small
\caption{Model specifications and single-request A100 latency.}
\label{tab:app_models}
\begin{tabular}{lcccc}
\toprule
\textbf{Model} & \textbf{Params} & \textbf{Precision} & \textbf{Serving Stack} & \textbf{Latency} \\
\midrule
Clef-flash & 9.0B & bfloat16 & PyTorch / HF & $\sim$120ms \\
StartLux & 4.1B & bfloat16 & PyTorch & $\sim$85ms \\
Open-Jev & 0.43B & float16 & PyTorch & $\sim$28ms \\
Qwen3.5-35B-A3B & 35B (3B act.) & bfloat16 & Transformers / vLLM & $\sim$9.3s \\
Qwen3-8B & 8.2B & bfloat16 & vLLM & $\sim$2.5s \\
\bottomrule
\end{tabular}
\end{table}

All experiments run on $8\times$ NVIDIA A100-80GB PCIe GPUs. $S_1$ forward passes and feature extraction use PyTorch 2.6 (CUDA 12.4); $S_2$ CoT decoding uses vLLM on PyTorch 2.7.1 (CUDA 12.6). EpiArb itself adds no training: all residual statistics are computed from cached $S_1$ states on CPU.

\paragraph{Where the evidence heads sit.}
Clef-9B's schema head (width $1024$) first builds one query per option from its span, its lexical embedding, and the question, and passes these queries through $2$ evidence-routing layers, each a $16$-head cross-attention over the projected backbone states followed by a feed-forward block.
The routed options are softmax-pooled into a field state, refined by $4$ decoder layers over the backbone states, and scored by Eq.~\eqref{eq:readout}.
EpiArb reads the $32$ cross-attention heads of the evidence-routing layers, enumerated layer-major ($0$--$15$ layer $1$, $16$--$31$ layer $2$).
StartLux and Open-Jev have no such head axis; for them the same residual formulas use a single option-level cosine channel (option state against question state for Open-Jev, final hidden state against each option's output row for StartLux), selected on validation.

\subsection{Controlled Multi-Field Benchmark Construction, Human Verification, and Public Splits}
\label{sec:app_data}

\begin{table}[ht]
\centering
\small
\caption{\textbf{Complete inventory of controlled multi-field decision probes constructed in this study.}
One multi-field episode (\texttt{state}) queries $1$--$2$ typed schema decisions simultaneously in a single $S_1$ forward pass.
All four splits ($950$ episodes, $1{,}455$ questions) were generated deterministically without LLMs; labels are exact by construction, with author generator audits and rendered-episode spot-checks.}
\label{tab:app_synthetic_suite}
\setlength{\tabcolsep}{3.4pt}
\begin{tabular}{llrrllp{5.5cm}}
\toprule
\textbf{Split} & \textbf{Paper Role} & \textbf{Eps.} & \textbf{Qs ($N$)} & \textbf{Seed} & \textbf{Schemas} & \textbf{Task Families (Questions) \& Diagnostic Purpose} \\
\midrule
\texttt{val} & Fit / val & 500 & 824 & 43 & \texttt{noul/choice/score} & Refund ($236$), security ($170$), triage ($134$), order logic ($108$), count ($92$), verify-math ($84$); $50/50$ split ($412/412$) for $\mathcal{V}$ and $(\tau_{\mathrm{local}},\mu,\sigma)$ \\
\midrule
\texttt{Hard} & Primary test & 150 & 240 & 20261006 & \texttt{noul/choice/score} & Refund chain ($80$), security policy ($62$), SLA route ($38$), verify-math ($36$), count logic ($24$); OOD conjunctions \& date shifts \\
\texttt{Hard\_v2} & Replication & 150 & 241 & 20261115 & \texttt{noul/choice/score} & Refund chain ($74$), SLA route ($64$), security policy ($44$), verify-math ($34$), count logic ($25$); ID offset $300\text{k}$, zero overlap \\
\texttt{Chain} & Depth probe & 150 & 150 & 20261008 & \texttt{choice} ($4$-way) & Chained arithmetic ($76$) and conditional ledger filter-sum ($74$); tests single-pass sequential compute bottleneck \\
\midrule
\multicolumn{2}{l}{\textbf{Test Subtotal}} & \textbf{450} & \textbf{631} & --- & Multi-field & Frozen test (\texttt{Hard}+\texttt{Hard\_v2}+\texttt{Chain}; Tables~\ref{tab:main_results}--\ref{tab:knockout}, \ref{tab:cross_model}--\ref{tab:residual_ctrl_ablation}) \\
\multicolumn{2}{l}{\textbf{Full Suite Total}} & \textbf{950} & \textbf{1,455} & --- & Multi-field & All $4$ splits; $100\%$ programmatic labels \\
\bottomrule
\end{tabular}
\end{table}

\paragraph{Why a custom multi-field suite is necessary.}
Standard public classification and QA benchmarks present one question with homogeneous flat options per forward pass, and frequently contain $1\%$--$3\%$ crowdsourced annotation noise.
For non-autoregressive decision models such as Clef-flash, StartLux, and Open-Jev, production inference evaluates \emph{multi-field typed schemas} in a single forward pass---for example, simultaneously judging boolean escalation (\texttt{noul}) and ordinal incident severity (\texttt{score}) over the same ticket narrative---where global pooling can preserve one field while washing out localized disqualifiers on another.
Moreover, because Tier-2 local arbitration flips a compact subset of high-conflict predictions without LLM calls, even a handful of mislabeled test items would confound \texttt{heal}-versus-\texttt{harm} safety accounting.
Our suite resolves both issues while pairing every mechanistic finding with six external public benchmarks (Table~\ref{tab:public_benchmarks}).

\paragraph{Three-stage construction and audit pipeline.}
Table~\ref{tab:app_synthetic_suite} summarizes all four splits. Cascade controllers are fit on held-out validation only (no System~1 fine-tuning on these probes).
\begin{enumerate}
    \item \textbf{Stage 1: Predicate-first deterministic synthesis (zero LLM generation).} Each generator first samples structured latent variables (e.g., purchase/evaluation dates, credential-leak booleans, SLA tier/product pairs, or multi-step ledger entries) under a fixed random seed, evaluates exact Python boolean/arithmetic rules to assign discrete ground-truth \texttt{labels}, and only then renders the natural-language \texttt{state} and randomly permuted \texttt{criteria} option orderings. Relative to \texttt{val}, \texttt{Hard} and \texttt{Hard\_v2} introduce deliberate distribution shifts: $3$-condition conjunctive refund rules with EU (\texttt{DD/MM/YYYY}), US (\texttt{MM-DD-YYYY}), and worded dates; $4$-predicate security runbooks mixing active threats with benign phishing-only distractors; an unseen SLA routing matrix; $2$--$3$ step arithmetic verification with near-miss distractors ($\pm 1,\pm 2,\pm 10\%$); and two-hop category-plus-flag counting. \texttt{Chain} constructs $4$--$5$ step running totals (\texttt{chain\_arith}) and conditional table aggregations (\texttt{filter\_sum}) whose three distractors equal intermediate partial results, sign-flipped final steps, or unfiltered sums (chance $=25\%$). (Generators also emit optional soft reference distributions for standalone proper-score checks; all cascade experiments evaluate strictly against exact discrete \texttt{labels}.)
    \item \textbf{Stage 2: Generator audit and rendered-episode spot-checks.} Labels follow the Stage~1 predicate evaluator. Authors audited each family generator against two criteria---(i)~\emph{text-to-label entailment} of the rendering template under the stated policy, and (ii)~\emph{schema and distractor validity} (mutually exclusive options; no accidental arithmetic collisions)---and spot-checked rendered episodes across all four splits for template bugs.
    \item \textbf{Stage 3: Strict fit/test isolation.} Validation split \texttt{val} ($824$ questions) is partitioned $50/50$ ($412/412$) to select head ensemble $\mathcal{V}$ via error-AUROC and fit $(\tau_{\mathrm{local}},\mu,\sigma)$, plus $10$ random A/B splits for stability analysis (Table~\ref{tab:residual_ctrl_ablation}). All parameters are frozen once before evaluating \texttt{Hard}, \texttt{Hard\_v2}, and \texttt{Chain}.
\end{enumerate}

\begin{table}[ht]
\centering
\small
\caption{\textbf{Representative multi-field episodes} from \texttt{Hard} (joint \texttt{noul} + \texttt{score} compliance with a benign phishing distractor) and \texttt{Chain} ($4$-way sequential filter-sum with partial-computation distractors).}
\label{tab:app_episode_example}
\begin{tabular}{lp{13.4cm}}
\toprule
\textbf{Episode ID} & \texttt{hard\_security\_policy\_200001} (\texttt{Hard}; $2$ joint questions in $1$ forward pass) \\
\textbf{State ($x$)} & \emph{``Security runbook R-7: escalate to SecOps iff any of (1) credential/secret exposure, (2) unauthorized admin/privilege change, (3) active ransomware/encryption. Phishing reports WITHOUT credential use are NOT escalate. Ticket narrative: backup volumes were wiped and a ransom email arrived $\mid$ user forwarded a suspicious email but no credentials were entered.''} \\
\textbf{Questions \& Gold} & $q_1$ (\texttt{noul}, \texttt{escalate\_security}): \emph{``Escalate to SecOps under runbook R-7?''} $\to$ \textbf{Gold:} \texttt{true} \newline
$q_2$ (\texttt{score}, \texttt{severity}, criteria \texttt{[None, Low, Medium, Critical]}): \emph{``Incident severity under R-7''} $\to$ \textbf{Gold:} \texttt{3} (\texttt{Critical}) \\
\midrule
\textbf{Episode ID} & \texttt{filter\_sum\_320002} (\texttt{Chain}; sequential conditional aggregation) \\
\textbf{State ($x$)} & \emph{``Ledger: sales:14, sales:7, sales:18, tech:14, billing:11, billing:16, tech:18, sales:9, billing:1. Sum amounts whose category=sales AND amount>=6. Ignore other rows.''} \\
\textbf{Question \& Gold} & $q_1$ (\texttt{choice}, \texttt{value}, options \texttt{\{108, 4, 59, 48\}}---unfiltered total, match count, perturbed sum, exact sum): \emph{``Sum of sales rows with amount>=6?''} $\to$ \textbf{Gold:} \texttt{48} \\
\bottomrule
\end{tabular}
\end{table}

\paragraph{External public benchmarks.}
Public suites are normalized into single-pass typed schema records with deterministic disjoint training, validation ($300$--$400$ items), and test ($300$--$600$ items) splits.
For each benchmark, the top $M{=}8$ directional cosine heads, $z$-score routing parameters, and $q_{80}$ local threshold are fit exclusively on the validation split and evaluated once on the test split; learned baselines (LR-bb, LR-wb, hidden-state probe) are trained on the same validation split.
Validation-side $\Delta_{\mathrm{inv}}$ (Table~\ref{tab:applicability}) is computed by $2$-fold cross-fitting inside the validation split ($10$ seeds): heads and $z$-scores are fit on one half and gain inversion is measured on the other, so no test item is used. The three primary probes share one validation split and hence one value.

\subsection{Hyperparameters and Synthetic Wash-Out Model}
\label{sec:app_settings}

Primary \texttt{Hard}/\texttt{Chain} experiments freeze:
\begin{itemize}
    \item \textbf{Selected heads ($\mathcal{V}$):} $M{=}8$ of the $32$ Clef-9B evidence-routing heads, $\{4,7,11,13,17,21,22,26\}$ (layer-major), with validation error-AUROC polarity on $\texttt{cos}$.
    \item \textbf{Local healing:} $r_{\mathrm{mass}}(x)\ge\tau_{\mathrm{local}}$ with $\tau_{\mathrm{local}}=0.7108$ (validation $q_{80}$) and $\hat{y}_{\mathrm{evi}}\neq\hat{y}_{\mathrm{joint}}$.
    \item \textbf{Escalation risk:} $R(x)=z(g(x))+z(-m_{\mathrm{evi}}(x))$, escalating top $K=\mathrm{round}(0.20\cdot|\mathcal{B}|)$ unhealed queries.
    \item \textbf{System~2 decoding:} greedy decoding ($T{=}0.0$, up to $768$ tokens) with structured \texttt{FINAL: <option\_id>} extraction ($98.7\%$ raw parse rate).
    \item \textbf{Mode gate:} enable local healing only if validation heals $\ge 2\times$ harms; otherwise Route only.
\end{itemize}

\paragraph{Synthetic wash-out model (\S\ref{sec:method_analysis}).}
Each seed draws $n{=}5{,}000$ binary items. A regime is sampled as noise with probability $p_{\mathrm{noise}}{=}0.15$, wash-out with probability $p_{\mathrm{wash}}\in\{0,0.05,0.1,0.2,0.35,0.5,0.65\}$, and easy otherwise.
Local evidence for each option is $L_{\mathrm{gold}}\sim\mathcal{N}(+\mu_L,0.35^2)$, $L_{\mathrm{prior}}\sim\mathcal{N}(-\mu_L,0.35^2)$ with $\mu_L{=}2.2$ on easy and wash-out items and $0.35$ on noise items; the global prior is $G_{\mathrm{prior}}\sim\mathcal{N}(+2,0.35^2)$, $G_{\mathrm{gold}}\sim\mathcal{N}(-2,0.35^2)$.
Readout logits are $z=(1{-}\alpha)L+\alpha G+\mathcal{N}(0,0.12^2)$ with $\alpha{=}0.9$ on wash-out items and $\alpha{=}0$ otherwise.
Head scores are $s=L+\epsilon$ with $\epsilon\sim\mathcal{N}(0,0.15^2)$ on easy and wash-out items and $\mathcal{N}(0,1.35^2)$ on noise items; this larger head noise on noise items is a modeling assumption encoding that heads are no more certain than the readout when local evidence is weak.
$S_2$ is an oracle, so escalate gain equals $S_1$ error. $M$ and $R$ are computed exactly as in \S\ref{sec:method_epiarb}, with $z$-scores fit per batch; we report pairwise gain-inversion rates averaged over $20$ seeds.

\subsection{Extended External Public Baselines and Cross-Backbone Transfer}
\label{sec:app_public_ext}

\begin{table}[ht]
\centering
\small
\caption{\textbf{Cross-backbone public benchmark $\Delta$ vs.\ Logit Margin @$20\%$ (pp)} (\texttt{Route}~/~\texttt{L+R}).}
\label{tab:app_public_multimodel}
\begin{tabular}{lccc}
\toprule
\textbf{Benchmark} & \textbf{Clef-flash 9B} & \textbf{StartLux 4B ($S_1$ Acc)} & \textbf{Open-Jev 0.4B ($S_1$ Acc)} \\
\midrule
PhishNChips & $\mathbf{+1.83 / +2.33}$ & $+0.00 / +0.17$ ($91.0\%$) & $+0.50 / +0.50$ ($50.0\%$) \\
RAGTruth & $\mathbf{+1.80 / +2.80}$ & $+0.00 / +0.00$ ($87.4\%$) & $\mathbf{+5.20 / +5.00}$ ($49.6\%$) \\
Clickbait & $\mathbf{+1.60 / +1.60}$ & $-0.20 / -0.20$ ($61.4\%$) & $+0.20 / +0.40$ ($50.0\%$) \\
\bottomrule
\end{tabular}
\end{table}

Table~\ref{tab:app_public_multimodel} reports cross-JEV transfer on the wash-out-prone public suites.
On low-capacity Open-Jev-0.4B, RAGTruth Residual Route yields $+5.20$pp over Margin ($55.8\%$ vs.\ $50.6\%$).
StartLux-4B is near-saturated on PhishNChips ($91.0\%$) and RAGTruth ($87.4\%$), leaving negligible cascade headroom over Qwen3-8B.

\subsection{Cross-Backbone Transfer and Design Ablations}
\label{sec:app_transfer_ablation}

\begin{table}[ht]
\centering
\small
\caption{\textbf{Cross-backbone transfer on \texttt{Hard}} (frozen residual definitions; mode gate fit on validation).
Vary-$S_1$ rows use Qwen3-8B escalate caches; vary-$S_2$ rows fix Clef-9B.
Gains are vs.\ Logit Margin @$20\%$.}
\label{tab:cross_model}
\setlength{\tabcolsep}{2.4pt}
\resizebox{\columnwidth}{!}{%
\begin{tabular}{llccc}
\toprule
\textbf{Axis} & \textbf{Backbone} & \textbf{Acc} & \textbf{Residual gain ($\Delta$pp)} & \textbf{Gated mode} \\
\midrule
\multirow{3}{*}{Vary $S_1$}
 & Clef-flash 9B & $S_1$ $84.58$ & Local $+2.50$; Route $+1.67$ & \texttt{L+R} \\
 & StartLux 4B & $S_1$ $94.17$ & Local $+2.50$ (security) & \texttt{Local} \\
 & Open-Jev 0.4B & $S_1$ $41.67$ & Route $+1.25$ (local unsafe) & \texttt{Route} \\
\midrule
\multirow{2}{*}{Vary $S_2$}
 & Qwen3.5-35B-A3B & $S_2$ $98.75$ & L+R $+2.92$ ($98.33$) & \texttt{L+R} \\
 & Qwen3-8B & $S_2$ $93.75$ & Route $+1.67$ ($95.00$) & \texttt{L+R} \\
\bottomrule
\end{tabular}%
}
\end{table}

\begin{table}[ht]
\centering
\small
\caption{\textbf{Design ablations and validation-fit stability on \texttt{Hard}.}
Left: local-tier $\Delta$pp vs.\ $S_1$ ($S_2$-free). Right: $10$ random validation A/B refits (routing uses the 8B cache).}
\label{tab:residual_ctrl_ablation}
\setlength{\tabcolsep}{2.6pt}
\resizebox{\columnwidth}{!}{%
\begin{tabular}{llc|lcc}
\toprule
\textbf{Axis} & \textbf{Variant} & \textbf{$\Delta$pp} & \textbf{10-seed refit} & \textbf{Local} & \textbf{Route} \\
\midrule
Heads & AUROC top-8 & $\mathbf{+2.50}$ & Mean $\pm$ std & $+2.50{\pm}0.00$ & $+1.37{\pm}0.28$ \\
 & Random-8 ($\times100$) & $-7.50$ & Min / Max & $[+2.50,+2.50]$ & $[+0.83,+1.67]$ \\
 & All heads & $-7.92$ & Gate pass & $10/10$ & $9/10$ \\
\cmidrule{1-6}
Signal & $r_{\mathrm{mass}}$ (ours) & $+2.50$ & \multicolumn{3}{l}{Mean head Jaccard vs.\ reference: $0.81$} \\
 & Binary Vote-$D$ & $+1.67$ & & & \\
\bottomrule
\end{tabular}%
}
\end{table}

\subsection{Learned Baselines and the Hidden-State Probe}
\label{sec:app_learned_router}

\paragraph{LR-bb / LR-wb.} Class-balanced logistic regression on standardized features (LR-bb: margin, max-probability, entropy, log option count; LR-wb: additionally $R$, $r_{\mathrm{mass}}$, $g$, $m_{\mathrm{evi}}$, and the binary disagreement flag), trained on validation to predict escalate gain (validation $S_2$ outcomes from the same $S_2$ cache), and used route-only: the top $20\%$ by predicted gain are escalated.

\paragraph{Hidden-state probe.} For every question we extract three $S_1$ states in the same forward pass: the pooled global state, the question state, and the per-field state feeding the readout. Their concatenation is standardized, reduced by PCA to $128$ dimensions, and fed to a class-balanced logistic regression whose $L_2$ strength is chosen by $5$-fold cross-validated AUROC on validation. We train two targets: escalate gain (reported in Tables~\ref{tab:main_results} and~\ref{tab:public_benchmarks}) and $S_1$ error (\citealp{azaria2023internal}).
The error-target probe gives \texttt{Hard} $95.00\%$, \texttt{Hard\_v2} $96.68\%$, \texttt{Chain} $62.67\%$, RAGTruth $73.6\%$, PhishNChips $83.0\%$, Clickbait $90.0\%$, BBH $69.8\%$, GSM8K $80.7\%$, and ANLI $57.2\%$, i.e.\ the same pattern as the gain target: at Margin on the shifted primary probes (residual cascade $+3.33$pp on \texttt{Hard}, $p{=}0.021$), strongest on i.i.d.\ public wash-out suites.
Test escalate-gain AUROC of the gain probe vs.\ Margin vs.\ residual $R$: \texttt{Hard} $0.934/0.903/0.957$, \texttt{Hard\_v2} $0.919/0.894/0.930$, RAGTruth $0.840/0.676/0.732$, PhishNChips $0.993/0.732/0.787$.

\paragraph{Route vs.\ L+R on \texttt{Hard}.} Under the primary protocol, residual Route alone reaches $97.08\%$ and the deployed L+R $98.33\%$ (Table~\ref{tab:knockout}).

\subsection{Cross-\texorpdfstring{$S_2$}{S2} Replication on \texorpdfstring{\texttt{Hard\_v2}}{Hard v2} and Feature Causal Knockout}
\label{sec:app_hard_v2}
\label{sec:app_causal}

\paragraph{Cross-$S_2$ replication (Qwen3-8B).}
Evaluating the frozen controller on \texttt{Hard\_v2} ($N{=}241$) with Qwen3-8B CoT ($S_2$ Acc.\ $95.02\%$) at a matched $20\%$ budget yields: Logit Margin $94.19\%$, Vote-$D$ Tri-Tier $95.02\%$, and Residual Tri-Tier $\mathbf{97.51\%}$ ($+2.49$pp vs.\ Vote-$D$, exact McNemar $b{=}6,c{=}0$, $p{=}0.031$, bootstrap CI $[+0.83,+4.56]$).

\paragraph{Causal feature ablation.}
On \texttt{Hard}, zeroing out the $\texttt{cos}$ features of the eight selected heads $\mathcal{V}$ eliminates $57\%$ of beneficial local heals (and induces $37$ false flips, dropping accuracy to $76.25\%$); injecting Gaussian noise removes $86\%$ of heals.
Because random alternate heads can overlap $5/7$ healed items while still suffering massive false-positive flips ($-7.50$pp net accuracy), we claim functional dependence on the selected ensemble's signed features rather than anatomical uniqueness of individual heads.

\end{document}